\documentclass{article} 
\usepackage{icml2025_conference_arxiv,times}

\usepackage{hyperref}
\usepackage{url}
\usepackage{amsfonts}
\usepackage{graphicx}
\usepackage{soul} 
\usepackage[T1]{fontenc}

\usepackage[export]{adjustbox}

\usepackage{array}
\newcolumntype{H}{>{\setbox0=\hbox\bgroup}c<{\egroup}@{}}
\usepackage{fancyvrb}
\newcommand{\Rset}{\mathbb{R}}
\newcommand{\argmin}{\mathrm{argmin}}
\newcommand{\Pb}{\mathbf{P}}
\newcommand{\zb}{\mathbf{z}}
\newcommand{\emb}{E}
\newcommand{\FSP}{\mathrm{SS}_\mathrm{NP}}

\newcommand{\MPK}{\mathrm{SS}_\mathrm{MPk}}
\newcommand{\MPtwo}{\mathrm{SS}_\mathrm{MP2}}

\newcommand{\MLPC}{\mathrm{SS}_\mathrm{MC}}
\newcommand{\HP}{\mathrm{PT}}
\newcommand{\HPSR}{\mathrm{PT}_\mathrm{SR}}
\newcommand{\TGSM}{\mathrm{TinyGSM}}
\newcommand{\SOFTSRVembedding}{SoftSRV embedding}
\newcommand{\SOFTSRVembeddings}{SoftSRV embeddings}

\usepackage{caption}
\usepackage{subcaption}

\title{SoftSRV: Learn to Generate Targeted Synthetic Data}

\author{Giulia DeSalvo, Jean-Fra\c{c}ois Kagy, Lazaros Karydas, Afshin Rostamizadeh, Sanjiv Kumar \\
Google Research \\
New York, NY 10011, USA \\
\texttt{\{giuliad, jfkagy, lkary, rostami, sanjivk\}@google.com} \\
}

\iclrfinalcopy 
\begin{document}

\maketitle

\begin{abstract}
We present a novel framework, SoftSRV, that is used to generate targeted synthetic fine-tuning data for improving task-specific model performance. Given a sample from a target distribution, our proposed framework uses a data-driven loss minimization approach to steer a frozen large language model (LLM) to generate synthetic sequences that are similar to those from the target distribution. SoftSRV provides a practical improvement over common prompt engineering approaches that rely on human-engineered prompt-templates, which can be idiosyncratic, labor-intensive to craft, and may need to be specialized per domain. We empirically evaluate our method against standard baselines guiding a large LLM to generate synthetic data to fine-tune a smaller language model on three different domains (coding, math, reasoning). We perform these evaluations without any particular specialization of the framework to each domain, emphasizing the generality of our approach. We find that SoftSRV improves upon typical prompt engineering approaches, generating targeted data that leads to fine-tuned models with significantly better task-specific performance. In addition, SoftSRV-generated data better matches the target distribution according to the {\sc mauve} similarity metric. 

\end{abstract}

\section{Introduction}

In recent years, pre-trained large language models have proven to be effective in generating synthetic natural language training data
\citep{gunasekar2023textbooks,li2023textbooks,eldan2023tinystories,mukherjee2023orca,mitra2023orca,abdin2024phi}.
This is particularly true when the synthetic data is used to pre-train or fine-tune smaller language models, enabling performances that rival models that are orders of magnitude larger \citep{liu2023tinygsm}. There are several motivations for generating and using synthetic training data; chief among them is the need to train models for domains where little natural high-quality text may be readily available or may be difficult to procure. 
  
In order to generate synthetic text, a significant amount of human-driven prompt engineering is invested into developing prompts that steer the generating LLM into producing high-quality text from a targeted domain while also encouraging sufficient diversity. This point was nicely summed up by the authors of the open-source synthetic text repository Cosmopedia \citep{cosmo}, when recounting their work to recreate a large synthetic dataset similar to the one generated to train Phi 1.5 \citep{li2023textbooks}:  

\begin{quote}
  ``Heads up: If you are anticipating tales about deploying large-scale generation tasks across hundreds of H100 GPUs, in reality most of the time for Cosmopedia was spent on meticulous prompt engineering.''
      \qquad-- \cite{cosmo}

\end{quote}
 
Furthermore, and especially in the case of generating fine-tuning data for targeted domains (e.g.,\ coding, math, customer service), this manual process may need to be repeated and refined per-domain, or even per sub-domain (e.g.,\ per coding language, math subject, service department).
Apart from the human engineering cost, these manual prompting approaches do not directly optimize a data-driven objective. Rather they depend on human-in-the-loop style feedback for manually adjusting the prompt templates, resulting in approaches that lack robust mechanisms for aligning the LLM's generated data with the desired distribution.


To address these issues, we propose an algorithmic framework, SoftSRV, that leverages trainable parameteric embeddings, rather than natural language prompts, to steer a pre-trained model towards generating text that most resembles the target distribution. These parametric embeddings are trained by minimizing a data-driven loss function using an autoencoder-like compression and reconstruction procedure. 
We restrict SoftSRV to train on only a small number of parameters, thereby using a relatively limited amount of compute. Additionally, SoftSRV requires essentially no human-in-the-loop prompt engineering, enabling the process to be readily deployed across many different domains. 

Parameteric embeddings allow for more expressive inputs to an LLM compared to natural language prompts since an embedding is not restricted to correspond to a particular sequence of discrete natural language tokens. This intuitive observation is formalized in \citet{petrovprompting}, which shows that in specific settings trained embeddings can induce an LMM to produce an exponential (in sequence length) number of text completions, while natural language prompts only allow for a linear number of completions. Although SoftSRV has some similarities to traditional prompt-tuning approaches, there are also several critical differences, which we discuss later in Section~\ref{sec:differences}.


The specific contributions presented in this work are:
\begin{itemize}
    \item We introduce the SoftSRV framework and provide several instances of parameterized contextual embeddings that can be leveraged within the framework.
    \item We demonstrate that these embeddings can be successfully trained and used to generate targeted synthetic text for fine-tuning task-specific downstream models.
    \item Our empirical evaluations demonstrate that models fine-tuned on SoftSRV-generated text admit superior performance compared to those fine-tuned on data generated by baseline prompt engineering approaches or on limited amount of non-synthetic data. We show results on coding, math, and reasoning benchmarks, using both in-domain and out-of-domain tasks to study generalization of the synthetic data.
    \item We measure the similarity of the generated data to the target distribution using the {\sc mauve} metric and observe that SoftSRV methods align most closely with the target distribution. 
\end{itemize}

\section{Proposed Approach}
\label{sec:proposed_approach}

\begin{figure*}
    \centering
    \includegraphics[scale=0.64]{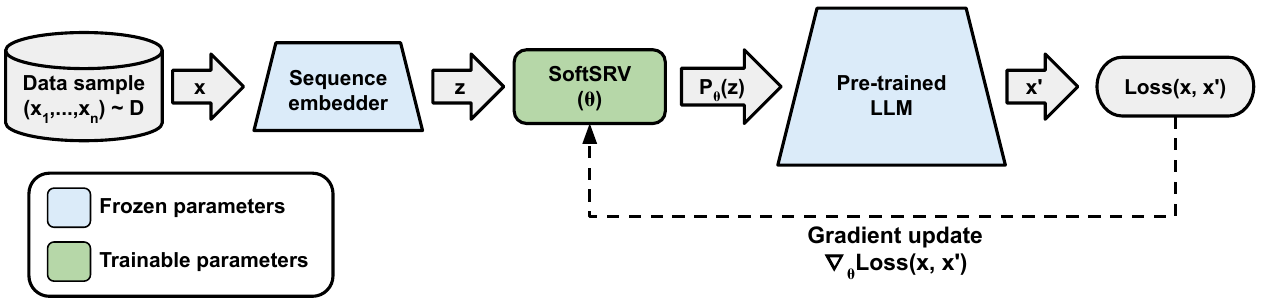}
    \caption{A diagram illustrating the training workflow of the SoftSRV framework. An example sequence $x$ is embedded into a dense vector $\zb$ via a (frozen) sequence encoder model. The SoftSRV model, parameterized by $\theta$ and conditioned on the embedding $\zb$, produces a \SOFTSRVembedding\ $\Pb_\theta(\zb)$. This is then fed to a (frozen) pre-trained LLM, which produces a synthetic example $x'$. Similar to autoencoder-based training, the gradient of a next-word-prediction ``reconstruction'' loss is computed and used to update the SoftSRV parameters.}
    \label{fig:framework}
\end{figure*}

In this section, we introduce the general SoftSRV framework along with a few specific parameterizations. First, we start with some basic notation and terminology.

Given a vocabulary $\mathcal{V}$ of textual tokens, let $\{x_1, \ldots, x_n\}$ denote a sample of $n$ text sequences, belonging to the set of all possible sequences $S^m$ of a maximum length $m$, drawn according to a fixed but unknown distribution $\mathcal{D}$. Although we are not able to directly sample additional sequences from $\mathcal{D}$, our goal is to synthesize sequences that could have plausibly been drawn according to $\mathcal{D}$.
We assume access to a (frozen) LLM, denoted $L: S^m \to S^m$, where we input and output sequences of equal fixed length $m$ for notational simplicity and without loss of generality. Furthermore, we explicitly decompose the LLM, $L = H \circ I$, where $I: S^m \to \Rset^{d \times m}$ represents the initial embedding layer that embeds each token of the input sequence to a $d$-dimensional dense vector, and $H: \Rset^{d \times m} \to S^m$ represents the remainder of the language model that maps the embedded tokens to the output sequence.
 For SoftSRV, we rely solely on the frozen model $H$ and we discard $I$ entirely. As explained below, we will replace $I$ by a parametrized family of embeddings.

The SoftSRV framework seeks to synthesize sequences similar to those drawn from $\mathcal{D}$ by training a dense embedding (or parameterized family of embeddings) $\Pb \in \Rset^{d \times t}$, with dimension $0<t<m$.  We hereafter refer to $\Pb $ as the \textit{\SOFTSRVembedding}. 
A successfully trained \SOFTSRVembedding, $\Pb$, should generate a sequence $x = H(\Pb)$, via the frozen model $H$, that has a high likelihood of occurring under the distribution $\mathcal{D}$. More generally, we can sample several different sequences from a learned \SOFTSRVembedding, $x, x', x'' \ldots \sim H(\Pb)$, by using randomized temperature-based decoding. 

It is a recognized problem within the field of synthetic data generation that generated datasets often lack diversity \citep{li2023textbooks,cosmo}.  Although temperature sampling alone does allow for some variability, it is important to be able to inject more diversity in the generated sample. To this end, we further increase the variety of generated text by introducing a \emph{contextual}  \SOFTSRVembedding, $\Pb(\cdot) : \Rset^{d_e} \to \Rset^{d \times t}$. A contextual  \SOFTSRVembedding\ can be conditioned with different context vectors $\zb \in \Rset^{d_e}$, during training and generation, to induce greater variations in the \SOFTSRVembedding. 


 

%
 Before introducing specific parametrized families of \SOFTSRVembeddings, we describe the SoftSRV training procedure which is common throughout and also illustrated in Figure~\ref{fig:framework}.  We first let $\theta$ denote the trainable parameters of the contextual \SOFTSRVembedding\ $\Pb_\theta(\cdot)$.   In addition to the sample of data ($x_1, \ldots, x_n$) and frozen LLM ($H$), we assume access to a frozen sequence embedding function $\emb(\cdot): S^m \to \Rset^{d_e}$ that will define our context vector  $\zb_i = \emb(x_i)$. During training, each training sequence is mapped to the context vector $\zb_i = \emb(x_i)$ and used to generate a conditioned \SOFTSRVembedding\ $\Pb_\theta(\zb_i)$, which is fed into the frozen LLM $H$ to produce a new sequence $x_i' \sim H(\Pb_\theta(\zb_i))$ using autoregressive next token generation.
A standard causal (next-word) prediction loss, denoted $\ell(\cdot, \cdot)$, is backpropagated through the network up to the \SOFTSRVembedding\ layer $\Pb_\theta(\cdot)$, and an SGD-style update is applied to $\theta$ using the gradient $\nabla_\theta \ell(x_i, x_i')$.
This loss can be thought of as a ``reconstruction'' error and the entire pipeline is akin to an auto-encoder. Viewing the pipeline through this lens, it is apparent that the sequence embedder $\emb(\cdot)$ should be sufficiently ``lossy'' in order to avoid making the learning problem trivial. This lossiness can be enforced by restricting the dimension $d_e$ of the embedding, for example.

Once the contextual \SOFTSRVembedding\ $\Pb_\theta(\cdot)$ has been trained, we can then generate synthetic data by passing $\Pb_\theta(\zb)$ to the frozen LLM as embedded input context for different choices of context vector $\zb$. A natural choice is to sample embeddings $(\zb_1, \ldots, \zb_n)$ derived from the data sample set $(x_1, \ldots, x_n)$.We now introduce a few specific SoftSRV parameterizations studied in this work.  

\subsection{SoftSRV Non-contextual Parameterization ($\FSP$)}
The simplest parameterization treats the $dt$ entries of a \SOFTSRVembedding\, $\Pb \in \Rset^{d \times t}$, directly as trainable parameters, i.e., $\theta =\Pb$ , resulting in the following objective:
\begin{equation}
  \argmin_\theta \sum_{i=1}^n \ell(H(\Pb), x_i) \,  \,.
\end{equation}
This parameterization is an instance of a \emph{non-contextual} \SOFTSRVembedding\, i.e., any context $\zb$ is ignored. Despite the lack of varying context, the synthesized output may still be diversified by using non-greedy (i.e., temperature sampling) decoding during LLM generation.

\subsection{SoftSRV Mixture Parameterization ($\MPK$)}

Here, we train $k$ ``basis'' \SOFTSRVembedding\ matrices and define the final \SOFTSRVembedding\, as a mixture of these bases.
More precisely, in this variant the parameter set is $\theta = \{\Pb_1,\ldots,\Pb_k,\phi\}$, where $\Pb_i \in \Rset^{d \times t}$ are the basis \SOFTSRVembeddings\, 
\begin{equation}
    \Pb_\theta(\zb) = \sum_{i=1}^k w_i \Pb_i \,, \quad (w_1, \ldots, w_k) = W_\phi(\zb),
\end{equation}
and $W_\phi(\cdot): \Rset^{d_e} \to \Rset^k$ is a learned softmax function with parameters $\phi \in \Rset^{d_w}$. The trained $\MPK$ embedding is then the SGD solution to $\argmin_\theta \sum_{i=1}^n \ell(H(\Pb_\theta(\emb(x_i)), x_i)$.

The intuition behind this formulation is for each learned basis embedding $\Pb_i$ to encode a different aspect (mode) of the target data distribution and have each training example $x_i$ approximated by a mixture of these modes (similar to the intuition behind mixture or topic models \citep{mclachlan1988mixture}). 



\subsection{SoftSRV MLP Concatenated Parameterization ($\MLPC$)}
Next, we consider a collection of $t$ small MLPs, whose output is concatenated to generate the final \SOFTSRVembedding . Let $F_{\phi_i} : \Rset^{d_e} \to \Rset^t$ denote the $i$th MLP with parameters $\phi_i$, and $\theta = \{\phi_1, \ldots, \phi_t\}$ denote the parameters for the collection of MLPs. Then, we define:
\begin{equation}
  \Pb_\theta(\zb) = \Big[ F_{\phi_1}(\zb), \ldots, F_{\phi_t}(\zb) \Big] \,,
\end{equation}
and the trained $\MLPC$ embedding is the SGD solution to $\argmin_\theta \sum_{i=1}^n \ell(H(\Pb_\theta(\emb(x_i)), x_i)$.

This parameterization is the most expressive that we consider, in that each embedding column (which can be thought of as an ``soft token'') is computed using a distinct non-linear transformation of the context vector $\zb$.

\subsection{Differences with Prompt-tuning}
\label{sec:differences}

Although SoftSRV and existing prompt-tuning methods \citep{lester2021power,li2021prefix} both train embeddings, we outline several crucial differences between the two approaches. (1) The application: prompt-tuning is generally used to prepend a  prompt embedding to an existing natural language prompt that is then fed to an LLM, with the goal of providing a higher quality response; in this work, our goal is to generate synthetic training data by feeding an LLM a trained embedding alone, i.e. there is no natural notion of an input/output pair in this application. 
(2) Contextual embedding: in standard prompt-tuning, it typically suffices to train a single prompt embedding that is prepended to different input prompts; in our setting, since we feed only the prompt embedding (and no natural language prompt) to the LLM we must do more than train a static embedding if we wish to induce significant variation. To that end, we introduce and train parameterized \emph{contextual} embeddings.
(3) The optimization procedure: in standard prompt-tuning the input/output sequence pairs (taken naturally from the application) can be used to train the prompt embedding; in the case of SoftSRV we train using an autoencoder-like compression and reconstruction scheme. For a discussion of the prompt-tuning literature, please see Appendix~\ref{app:soft_prompt_related_work}.


\section{Empirical Evaluation}

To empirically evaluate the SoftSRV framework, we consider a supervised fine-tuning setting where we fine-tune a small Gemma 2B model \citep{team2024gemma} using synthetic data generated by a large foundational decoder-only language model. To generate the data, we only consider methods that directly prompt the larger model and/or train a relatively small number of parameters (as in the case of SoftSRV). We do not assume the ability to conduct full fine-tuning of the large foundational model, as it is generally computationally prohibitive.

\subsection{Domains and Datasets}\label{datasets-subsection}

To demonstrate the generality of the proposed approach, we generate fine-tuning data for several disparate domains (coding, mathematics, reasoning) using the same exact pipeline with no particular specialization to any of the domains. 
 
  {\bf{Code -- MBPP \citep{austin2021program}}} consists of short Python programming exercises (metric: 3-shot pass@1).
  
  {\bf{Math -- GSM8K \citep{cobbe2021gsm8k}}} contains grade school math word problems  (metric: 5-shot pass@1).  
 
  {\bf{Reasoning -- BoolQ \citep{clark2019boolq}}} is a question answering dataset where each true/false question is paired with a passage (metric: accuracy).

The above benchmarks aim to cover a wide variety of tasks, each with varying degrees of complexity -- both in terms of solving the task and in terms of generating synthetic data. Please see Appendix~\ref{app:task_discussion} for a detailed discussion.

\subsection{Baselines}\label{hard-prompting-subsection}
Typical prompt engineering approaches for synthetic text generation involve manually creating  natural language  prompt templates that are then seeded with text from the desired target domain, usually taken from the training set. To give a simple illustrative example, a template could be: 
\begin{quote}
{\small{\texttt{Consider the following [article], write a summary of the topic suitable for a high-school audience}},}
\end{quote}
where the placeholder {\small{\texttt{[article]}}} would be replaced with example texts from training fold, producing several distinct prompts. 
In this study, we consider the following two natural language prompt template variants. 

The first, denoted simply as prompt template ($\HP$), uses a domain-specific template to generate a question followed by another domain specific template to generate answers (the detailed workflow is discussed in Subsection~\ref{pipeline}). 
In Appendix~\ref{app:hp}, we provide the exact templates used by the $\HP$ method. To design these templates, we undertook several iterations of prompt engineering and reported the result of the best performing method. In particular, we found that prompting for a "diverse" set of questions was crucial (a comparison plot is presented in Appendix~\ref{app:diversify}). 

The second approach, prompt template with self-refinement ($\HPSR$), similarly uses a template to generate questions but also iteratively conducts several rounds of self-critique to improve or accept the question \citep{madaan2024self}. Again, critique and refinement prompts are shown in Appendix~\ref{app:hp}.

We focus on comparing to natural language template prompting approaches as they are currently the most effective, general, and widely used approaches for practical synthetic data generation (see related work in Section~\ref{sec:related_work}).

\subsection{Empirical Evaluation Procedure}\label{pipeline}
\begin{figure*}
    \centering
    \includegraphics[scale=0.55]{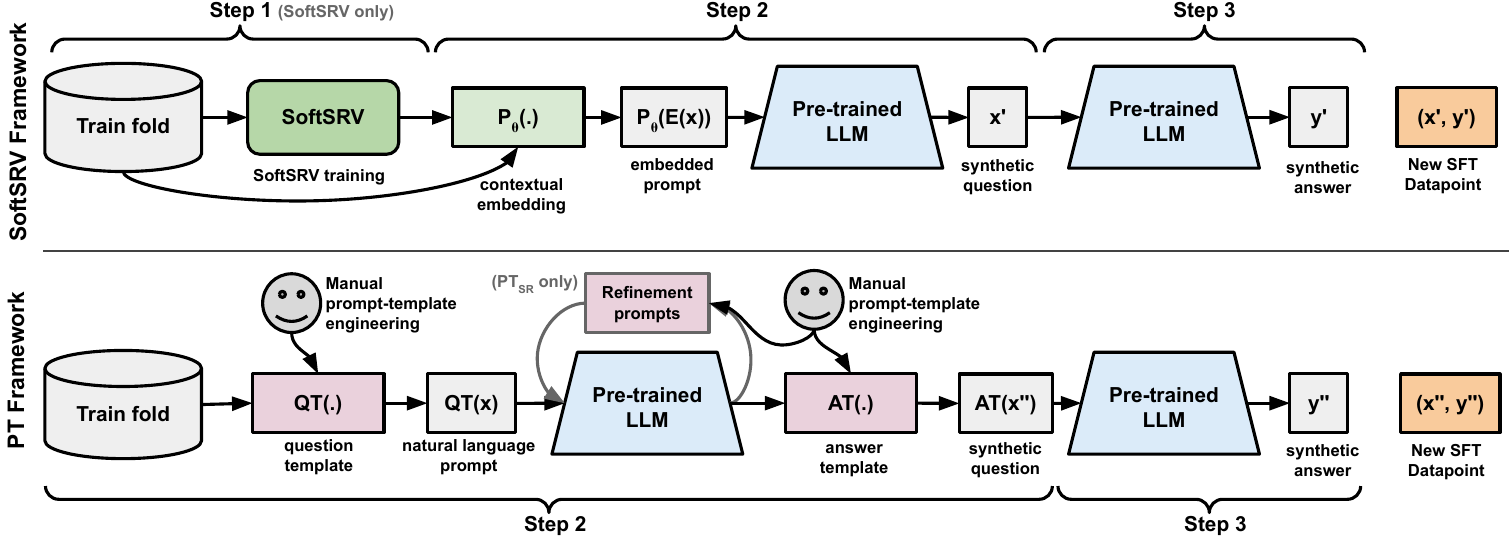}
    \caption{An illustration of the workflow for generating synthetic SFT data with SoftSRV (top panel) and natural language prompt templates (bottom panel); step numbering matches the discussion in Section~\ref{pipeline}. For SoftSRV, we use questions from the train set to (1) train a contextual embedding, (2) embed the same training sequences to serve as context vector to the trained embedding which is then fed to the LLM to generate synthetic questions, and (3) we generate answers by simply feeding the synthetic questions to an LLM. In the baseline prompt template framework, we have no training step (1) albeit there is offline human effort needed to generate the various prompt templates for each data domain; step (2) generates questions, using questions from the train set to fill the natural language template (optionally conduct rounds of refinement prompting), and (3) generates answers to these questions again using a template, but populated with the synthetic questions.}
    \label{fig:eval_framework}
\end{figure*}

Below, we describe the data generation and evaluation pipeline used for each benchmark dataset and each data generation approach.  Figure~\ref{fig:eval_framework} also provides a diagram, illustrating the data generation pipeline step used by SoftSRV and the prompt template baseline approaches.

{\bf Step 1. Train Parameters (SoftSRV only).} For the SoftSRV methods, we first train its parameters with a frozen large decoder-only LM backbone using the questions found in the training fold of the benchmark dataset, which serves as our sample from the target distribution.
We embed each question, $\zb_i = \emb(x_i)$, and run an Adam optimizer to minimize the causal next-word-prediction loss, $\argmin_\theta \sum_{i=1}^n \ell(H(\Pb_\theta(\zb_i), x_i)$, where $\Pb_\theta(\zb)$ is the contextual SoftSRV embedding and $H$ is the frozen LLM (post input embedding layer). The sequence embeddings, $\emb(\cdot)$, are set to be the average of token embeddings computed by a small off-the-shelf decoder-only LM. This simple choice for $\emb(\cdot)$ is used to both limit the amount of additional computation, but also to ensure the sequence embedding is somewhat lossy in order to make the reconstruction task (i.e. minimizing $\ell(H(\Pb_\theta(\emb(x_i)), x_i$) challenging). The simpler $\FSP$ method does not use this sequence embedding as it is not a contextual parameterization.

Since we are interested in a low-touch general framework, we avoid any domain-specific hyperparameter tuning for SoftSRV.  Specifically, for all SoftSRV variants and all benchmarks, the length of the prompt $t$ is fixed to be 128, the number of training steps was set to 20K, and the learning rate is fixed to $5\mathrm{e}{-6}$, which we found to be reasonable defaults. The $\MLPC$ method uses MLPs with 3 feed forward layers and 128 hidden dimensions. For the $\MPK$ variant, we primarily evaluate with $k=2$ to limit to the total number of parameters, although a partial exploration for other values of $k$ is presented in Appendix~\ref{app:mixture}.

{\bf Step 2. Generate Questions.} 
The main application of SoftSRV is in this step, where we generate new questions for the target fine-tune task.
Both the SoftSRV and the prompt template methods use all examples in the train data during this question generation phase. Specifically, SoftSRV uses examples $x_i$ from the train fold as context to generate a synthetic questions $x_i' \sim H(\Pb_\theta(\emb(x_i)))$, while $\HP$ and $\HPSR$ use $x_i$ to populate a template that is then passed to the LLM. Both approaches leverage temperature sampling.  We generate 50,000 questions for MBPP and GSM8K and 20,000 questions for BoolQ (see Appendix~\ref{app:examples} for examples). Appendix~\ref{app:postprocess} contains additional details on this step.

{\bf Step 3. Generate Answers.} 
After generating the questions, all methods essentially follow the same procedure to generate answers using an off-the-self LLM. The only difference being, in the case of SoftSRV, we pass the question directly to the LLM without any domain specific prompting to preserve its domain agnostic nature. In the case of the prompt template baselines, we use a domain specific template combined with the generated question to query the off-the-shelf LLM for an answer.
For all methods, once we have full (questions, answer) fine-tuning examples generated, we run a decontamination process to remove any examples that may have been inadvertently leaked to the pretrained LLM, as is standard practice (details provided in Appendix~\ref{app:decontaminate}).

{\bf Step 4. Fine-tune and Evaluate Downstream Model.} Finally, for all methods, we use the generated (question, answer) pairs to fine-tune the target 2B Gemma 2 model. We use a batch-size of 16 with sequence length 8192 and with a learning rate with linear warmup from 0 to 1e-6 over 100 steps, followed by a cosine annealing schedule.
We evaluate the performance of these fine-tuned models on the test fold of the respective benchmark using the metrics in Section~\ref{datasets-subsection}.

\subsection{ Comparing our Parameterizations of SoftSRV } \label{softsrv_ablation}
 
 \begin{table}
\caption{Downstream task performance of Gemma 2B models fine-tuned on the data generated by each of the three SoftSRV methods, evaluated at the 2K checkpoint. The number of non-synthetic examples used as seed data for the prompt templates and SoftSRV models is reported as $N_r$.  }
\centering
{
\begin{tabular}{l|Hl|lll}
task & metric & $N_r$ &  $\FSP$ & $\MPtwo$  & $\MLPC$\\
\hline
\hline
MBPP & pass@1 & 384 & 0.296	& 0.312 &    0.372   \\
GSM8K & accuracy & 7,473 & 0.413 & 0.401 &     0.463   \\
BoolQ  & accuracy & 9,427 & 0.806 & 0.825 &    0.831  \\
\hline
\end{tabular}
}\label{table:softsrvablation} 
\end{table}

We start our empirical analysis by comparing different SoftSRV parameterizations to identify the highest performing configuration.  Here, two salient questions are 1) Which parameterization most is effective and 2) whether it is essential to have a contextual \SOFTSRVembedding\ that leverages the context vector as opposed to a non-contextual \SOFTSRVembedding. In Table ~\ref{table:softsrvablation}, we present the performance of Gemma 2B fine-tuned on synthetic datasets generated by $\FSP$, $\MPtwo$  and $\MLPC$, respectively.  These results show that the non-contextual method, $\FSP$, generates synthetic fine-tuning data that results in lower performance than the other (contextual) parameterizations, indicating the effectiveness of contextual \SOFTSRVembedding. Further, comparing the performance of the $\MLPC$ and $\MPtwo$ methods, we observe that the more expressive parameterization of $\MLPC$ is generally beneficial. Consequently, we advocate for the use of the $\MLPC$ method in preference to other parameterized families.  We focus subsequent sections on comparisons with $\MLPC$.

\subsection{Comparing to Natural Language Templates} \label{downstream}

\begin{figure}
\begin{center}
    \includegraphics[scale=0.36]{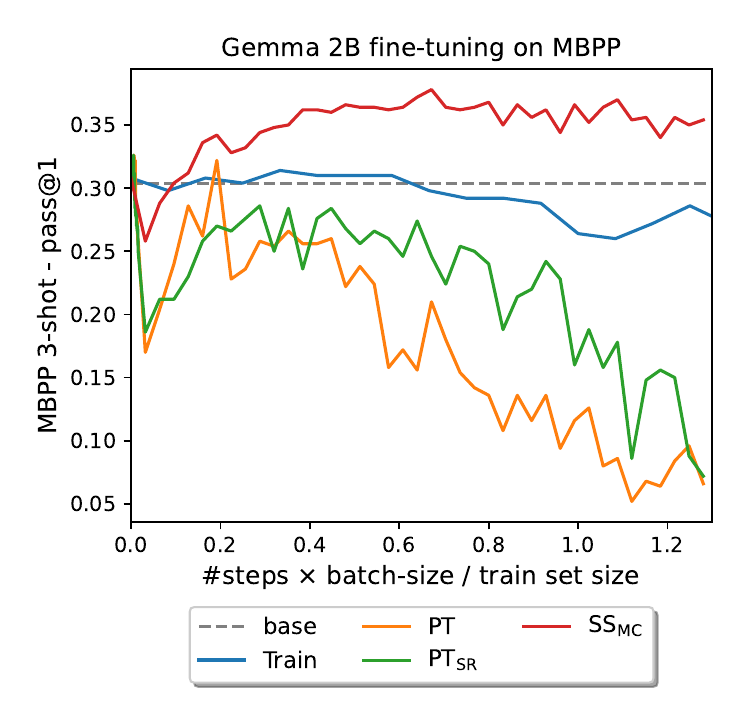}    
    \includegraphics[scale=0.36]{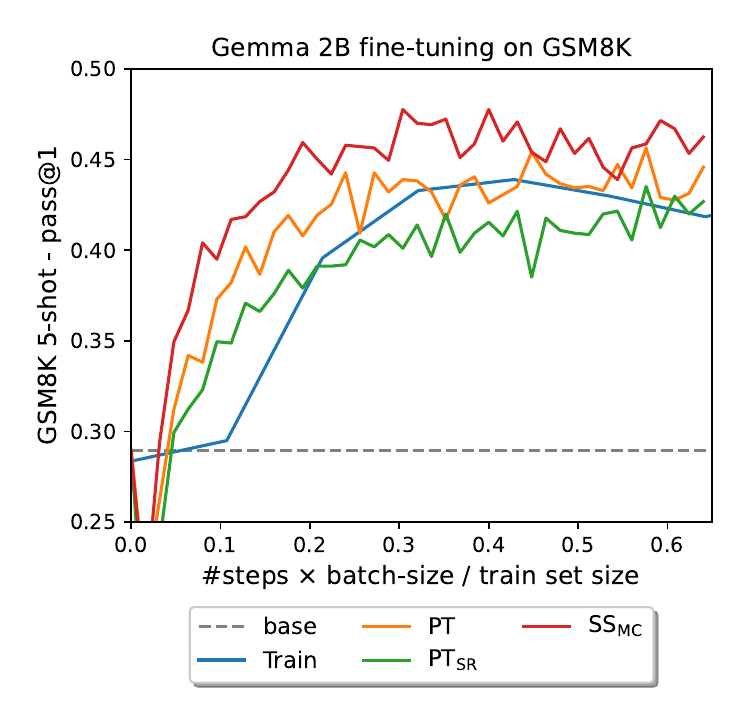}    
    \includegraphics[scale=0.36]{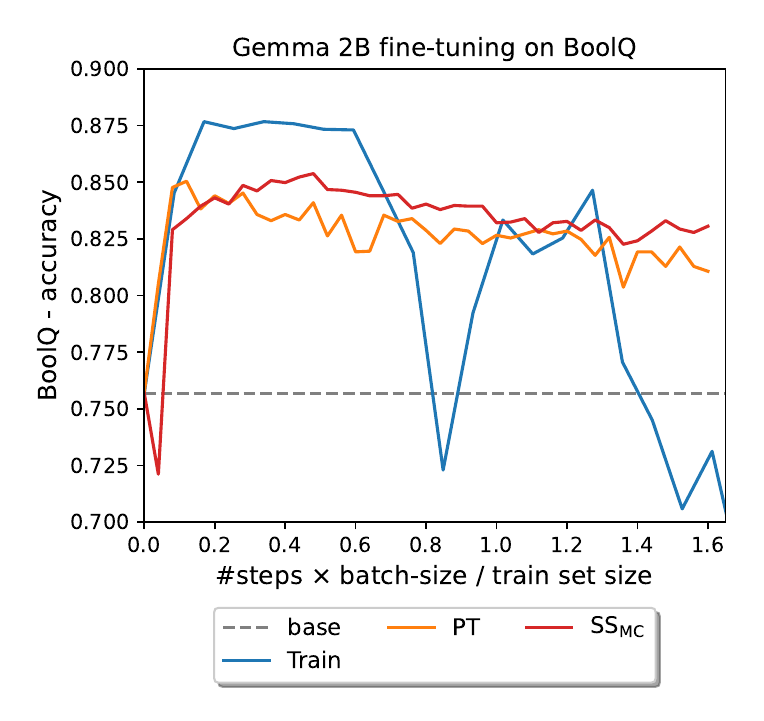} 
    \caption{Gemma 2 (2B) fine-tuning curves for the synthetically generated datasets as well as the non-synthetic training set.}
    \label{fig:gemma2b_curve}
\end{center}
\end{figure}


Here, we compare our best SoftSRV parameterization, $\MLPC$, to the two natural language prompt template methods described in Section~\ref{hard-prompting-subsection}. We first report the performance of Gemma 2B fine-tuned on the synthetic datasets generated by $\HP$, $\HPSR$, and $\MLPC$. Figures~\ref{fig:gemma2b_curve} plot the eval metrics for each dataset as a function of the number of fine-tuning steps times batch size normalized by training set size (essentially the training epoch modulo an additional constant factor due to sequence packing). We find that the model fine-tuned on $\MLPC$-generated data outperformed the models fine-tuned on data generated by the two prompt template methods.

Comparing the prompt template methods, the $\HP$ method outperforms the $\HPSR$ method on the GSM8K benchmark.  For the MBPP benchmark, $\HP$ initially attains a similar performance as that of $\HPSR$, but both methods start to degrade as a function of fine-tuning steps.  This may be due to a lack of diversity in the generated text, given the small set of 384 MBPP training examples used to seed the prompt templates. We do not report the results of the $\HPSR$ method for BoolQ as it appears to struggle to produce reasonable outputs. The repeated self-critiques of $\HPSR$ on the lengthy input passages seems to lead it astray from the original intention of the task, causing it to produce questions asking for open-ended discussion of a passage rather than targeted true/false questions.

We also emphasize that the training datasets from our benchmark domains contain a relatively small number of examples (see column $N_r$ of Table~\ref{table:softsrvablation}) and both prompt template methods and SoftSRV use this training data to generate tens of thousands of synthetic examples (see Step 2 in Section~\ref{pipeline} for exact number).  Yet, as shown by the results in this section, SoftSRV is more successful at leveraging this limited data.  Consequently, we argue that SoftSRV is especially well suited to the data scarce setting. 

Finally, as an additional comparison, we test $\MLPC$ generated data against the expertly curated GSM8K-specific synthetic dataset of \citet{liu2023tinygsm} (see detailed discussion in Section~\ref{tinygsm}).  Even though differences in the synthetic generation setup introduce several confounding factors, it is nonetheless encouraging to see that the $\MLPC$-generated data is comparable to this highly curated synthetic dataset.

\subsubsection{Out-of-Domain Benchmarks}
\begin{table}
\caption{Out-of-domain task performance of Gemma 2B models fine-tuned on $\HP$-generated data versus on $\MLPC$-generated data, evaluated at the 2K checkpoint.}
\centering
{
\begin{tabular}{lH|ll}
task & metric &   $\HP$  & $\MLPC$\\
\hline
\hline
Transcoder  & pass@1 & 0.141 & 0.369    \\
MAWPS & accuracy &  0.583  &   0.547	 \\
TriviaQA  & accuracy & 0.163 &  0.432 \\
\hline
\end{tabular}
}\label{table:out_domain_regular} 
\end{table}

The primary focus of this work has been on generating task-specific synthetic data and, thus far, we have evaluated fine-tuned models on held-out test sets from the task-specific distribution.
In this section, we measure the performance of the fine-tuned models on different, but thematically related, benchmarks to understand the generalizability of each data generation method to out-of-domain tasks. Here, we consider the following additional benchmarks: TriviaQA  \citep{joshi2017triviaqa} with a 5-shot prompt (accuracy) for reading comprehension, Transcoder \citep{sun2023transcoder} with a 3-shot prompt (pass@1) for coding, and MAWPS \citep{koncel2016mawps} with 0-shot (accuracy) for math.  Each of these metrics is tested on its corresponding model, e.g. for reasoning, we evaluate on TriviaQA  the model fine-tuned on the BoolQ-based synthetic datasets, etc.  Table~\ref{table:out_domain_regular} shows that the $\MLPC$ method outperforms  the $\HP$ template method on TriviaQA and Transcoder while exhibiting a closer performance on MAWPS.
 
Considering both in-domain results of Figures~\ref{fig:gemma2b_curve} and the aforementioned out-domain benchmarks, we find that, on the whole, the $\MLPC$ method exhibits superior performance compared to the $\HP$ template method.

\subsection{Comparing to the Original Training Data}\label{sec:comptrain}

Having identified $\MLPC$ as the most promising synthetic data generation method from those we considered, we now proceed to evaluate its performance relative to the original (non-synthetic) training data from the benchmarks. Figures~\ref{fig:gemma2b_curve} show that the model fine-tuned on the $\MLPC$ generated data outperforms the model fine-tuned on the non-synthetic train data for MBPP and GSM8K, indicating that, given enough of it, synthetic data can outperform even non-synthetic data. However, the same observation does not hold for BoolQ. The training set curve on BoolQ admits high variance, but it attains a higher accuracy overall. As discussed in Appendix~\ref{app:task_discussion}, we expect generating questions for the BoolQ dataset to be more difficult both due to the broad range of question topics and also due to the relatively long context-length needed to form a good question.

\subsubsection{Supplementing a post-training dataset}

In the previous evaluations, we considered a setting where we fine-tune using either the synthetically generated data or the original training data. 
Here, we investigate augmenting an existing general SFT dataset to evaluate the effectiveness of our datasets when combined with a mixture of data sources. Specifically, we compare the performance of a 2B Gemma 2 model fine-tuned on the Gemma 2 post-training dataset \citep{team2024gemma} augmented with the synthetic $\MLPC$ data versus augmented with the original training data. For all experiments, we fixed a priori the mixture source ratio to 80\% for the post-training dataset and 20\% for the added set (i.e. either $\MLPC$ data or training data). All other parameters remain consistent with previous sections except here the  models are trained for 10K steps as the amount of data is greater. 

Table~\ref{table:sft_table} shows that on MBPP and GSM8K, the model fine-tuned on the mixture of Gemma 2 post-training data with $\MLPC$ generated data outperforms the model trained on the mixture of the post-training with the train data. On BoolQ, similar to our previous findings (see first paragraph of Section~\ref{sec:comptrain}), the $\MLPC$-based model underperforms compared to the model using the train data. 


\begin{table}
\caption{ Downstream task performance of Gemma 2B models fine-tuned on a mixture of Gemma 2 post-training (G2) with $\MLPC$-generated data versus with the train data, evaluated at the 10K checkpoint. }
\centering
{
\begin{tabular}{lH|cc}
task & metric &   G2 + Train  & G2 + $\MLPC$\\
\hline
\hline
MBPP  & pass@1 &  0.320  &  0.356   \\
GSM8K & accuracy & 0.423  &  0.539  \\
BoolQ  & accuracy & 0.867  & 0.840  \\
\hline
\end{tabular}
}\label{table:sft_table} 
\end{table}

\subsection{Data Scaling}
\begin{figure}
\centering
  \includegraphics[scale=0.43]{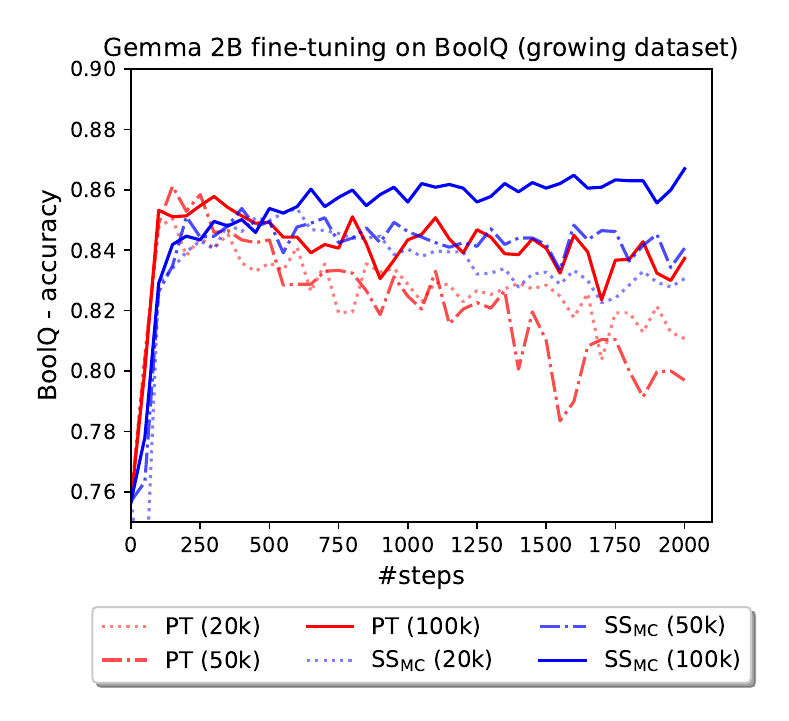}
  \vspace{-0.2cm}
  \caption{We compare the BoolQ performance of Gemma 2B fine-tuned on data generated by  $\HP$ and $\FSP$ as the number of generated examples increases.} 
  \label{fig:datascaling}
\end{figure}

Next, we analyze the effects of varying the number of synthetic examples generated when fine-tuning only on synthetic examples. Specifically, we increase the number of generated examples from 20K, 50K, to 100K, testing both the $\MLPC$ and $\HP$ approaches for the BoolQ benchmark, given this benchmark appears the most challenging in terms of generating effective synthetic questions. Figure~\ref{fig:datascaling} shows that the performance of the model fine-tuned on $\HP$ stagnates between 20K and 50K synthetic examples, and only shows improvement when the number of generated examples reaches 100K.  In contrast, the model fine-tuned on $\MLPC$ steadily improves as the generated dataset size increases. Going from 20K to 100K examples,  $\MLPC$  performance increases at a 1.8 times faster rate relative to $\HP$ with respect to their max values.  In particular, the $\HP$ method with 100K generated examples admits a comparable performance to that of the $\MLPC$ method with 50K generated examples. 

\subsection{Distribution Matching Analysis}


In addition to our evaluations of the fine-tuned model's downstream performance, we investigate the capacity of SoftSRV and the prompt template baselines to generate data samples that can match a target distribution of text. To assess the proximity between the generated and target distributions, we compute {\sc mauve} scores as in \citet{pillutla2021mauvemeasuringgapneural}, which can be viewed as a scalar summary of the divergence between the textual output of a generative model and a reference distribution. The {\sc mauve} score is able to simultaneously measure both the model's ability to avoid generating text outside the support of the target distribution (Type I error) and the ability to generate text with a large coverage of the target distribution support (Type II error). The method essentially computes a quantized distribution for the generated and target distribution and measures their KL divergence, producing  a normalized score between 0 and 1, where 1 indicates the two distributions are maximally similar (for further details see Appendix~\ref{app:mauve}). 

\begin{table} 
    \centering
    \caption{The {\sc mauve} scores of synthetic datasets computed with respect to the non-synthetic test fold of each dataset. 
    }
    \vspace{0.1in}
    \begin{tabular}{|l|c|c|c|} 
        \hline
         & MBPP & GSM8K & BoolQ \\
        \hline
        $\HP$ & 0.463 & 0.914 & 0.784 \\
        $\HPSR$ & 0.397 & 0.865 & -- \\
        $\MLPC$  &  0.477   &   0.991 & 0.995  \\
        \hline
        Train &  0.963 & 0.998 & 0.999 \\
        \hline
    \end{tabular}
    \label{mauve-table}
\end{table}

In Table~\ref{mauve-table}, we report the {\sc mauve} scores for the template prompt methods and for $\MLPC$ on the MBPP, GSM8K and BoolQ domains, measuring the distance to questions in the test fold (distance to question in the train fold are shown in Appendix~\ref{app:mauve}).
In all cases, we find that $\MLPC$ can synthesize text that is closer to a sample from the target distribution than the prompt template approaches. 
Notably, the $\MLPC$ method achieves a high {\sc mauve} score that is even close to that of the train set on GSM8K and BoolQ, where it appears the additional flexibility afforded by its parameterization allows a very high-fidelity match to the target distribution.  
While the prompt template-based datasets generally have lower {\sc mauve} scores than $\MLPC$, the $\HP$ variant is able to achieve a relatively high score on GSM8K. 
In the case of the MBPP dataset, we measure very low similarity scores for $\HPSR$; we conjecture this may be due to the several rounds of rewriting, which results in straying further from the original seed question.


Even though the {\sc mauve} score is not a direct indicator of downstream fine-tuning performance (see our experiments in Section~\ref{downstream} for performance comparisons), it is an additional signal to provide further support for SoftSRV as a high-fidelity approach for text generation across domains.

\section{Related Work}\label{sec:related_work}
As mentioned in the introduction, there is a significant recent body of work demonstrating the effectiveness of using a LLM to generate synthetic training data for a smaller model.

In terms for generating pre-training data, the collection of ``Textbooks Are All You Need'' white-papers outlines the process of training the Phi series of small LMs, using carefully curated prompt templates and seed data sources, and shows the large boost in quality that synthetic data can provide \citep{gunasekar2023textbooks,li2023textbooks,abdin2024phi}. The Cosmopedia project \citep{cosmo} conducts a similar study, while 
open-sourcing the prompts and generated data.
\citet{li2024synthetic} construct a pretraining set almost ``from scratch'' using a diverse set of prompt templates by using taxonomies of fields/sub-fields/disciplines within an area of expertise. 
Apart from focusing on pre-training rather than fine-tuning, these works require a non-trivial amount of human effort for building and/or curating prompt templates for the generating LLM, which our effort seeks to minimize.


\citet{mukherjee2023orca,mitra2023orca} focus on building synthetic data for better instruction tuning. In particular, they start with the FLAN-V2 instruction tuning dataset and ask a LLM to expand on the terse responses in different verbose styles (specified by so-called ``system instructions'') to introduce variation in presentation and approach. Although, shown to be quite effective across a broad array of reasoning tasks, in our setting we wish to generate data focused on a specific target task, likely requiring us to curate a set of bespoke ``system instructions'' for each task.

Several works build fine-tuning data for specific domains, such as coding \citep{haluptzoklanguage,luowizardcoder} or mathematics \citep{yumetamath}. Although quite successful, these approaches leverage specific qualities of the target domain, for example, using a code interpreter to check correctness of generated code or using the fact that math problems contain numerical quantities that can be masked or manipulated to create variations of the original question.


Finally, \citet{lee2024llm2llm} recently proposed an adaptive procedure where a LLM is used to generate targeted fine-tuning data for a small model based on examples that the small model has made mistakes on. The LLM is prompted to rewrite variants of these questions using specialized per-domain prompt templates. Extending SoftSRV to an adaptive setting is a potentially promising future line of work.

\section{Conclusion}

In this work, we established the effectiveness of the SoftSRV framework for generating targeted synthetic fine-tuning data. 
We deploy the same SoftSRV pipeline across math, coding, and reasoning tasks, finding in each case that SoftSRV generates fine-tuning data that provides strong downstream performance with no manual prompt engineering or per-domain specialization needed. A potentially fruitful direction for future work is to adaptively select or generate context vectors $\zb$, in order to generate even more impactful synthetic data.







\bibliography{softsrv}

\begin{thebibliography}{34}
\providecommand{\natexlab}[1]{#1}
\providecommand{\url}[1]{\texttt{#1}}
\expandafter\ifx\csname urlstyle\endcsname\relax
  \providecommand{\doi}[1]{doi: #1}\else
  \providecommand{\doi}{doi: \begingroup \urlstyle{rm}\Url}\fi

\bibitem[Abdin et~al.(2024)Abdin, Jacobs, Awan, Aneja, Awadallah, Awadalla,
  Bach, Bahree, Bakhtiari, Behl, et~al.]{abdin2024phi}
Abdin, M., Jacobs, S.~A., Awan, A.~A., Aneja, J., Awadallah, A., Awadalla, H.,
  Bach, N., Bahree, A., Bakhtiari, A., Behl, H., et~al.
\newblock Phi-3 technical report: A highly capable language model locally on
  your phone.
\newblock \emph{arXiv preprint arXiv:2404.14219}, 2024.

\bibitem[Austin et~al.(2021)Austin, Odena, Nye, Bosma, Michalewski, Dohan,
  Jiang, Cai, Terry, Le, et~al.]{austin2021program}
Austin, J., Odena, A., Nye, M., Bosma, M., Michalewski, H., Dohan, D., Jiang,
  E., Cai, C., Terry, M., Le, Q., et~al.
\newblock Program synthesis with large language models.
\newblock \emph{arXiv preprint arXiv:2108.07732}, 2021.

\bibitem[Ben~Allal et~al.(2024)Ben~Allal, Lozhkov, and van Strien]{cosmo}
Ben~Allal, L., Lozhkov, A., and van Strien, D.
\newblock Cosmopedia: how to create large-scale synthetic data for
  pre-training.
\newblock \url{https://huggingface.co/blog/cosmopedia}, 2024.

\bibitem[Brown et~al.(2020)Brown, Mann, Ryder, Subbiah, Kaplan, Dhariwal,
  Neelakantan, Shyam, Sastry, Askell, et~al.]{Brown2020}
Brown, T., Mann, B., Ryder, N., Subbiah, M., Kaplan, J.~D., Dhariwal, P.,
  Neelakantan, A., Shyam, P., Sastry, G., Askell, A., et~al.
\newblock Language models are few-shot learners.
\newblock In \emph{Neural Information Processing Systems}, 2020.

\bibitem[Clark et~al.(2019)Clark, Lee, Chang, Kwiatkowski, Collins, and
  Toutanova]{clark2019boolq}
Clark, C., Lee, K., Chang, M.-W., Kwiatkowski, T., Collins, M., and Toutanova,
  K.
\newblock {B}ool{Q}: Exploring the surprising difficulty of natural yes/no
  questions.
\newblock In \emph{{North {A}merican Chapter of the Association for
  Computational Linguistics: Human Language Technologies}}, 2019.

\bibitem[Cobbe et~al.(2021)Cobbe, Kosaraju, Bavarian, Chen, Jun, Kaiser,
  Plappert, Tworek, Hilton, Nakano, et~al.]{cobbe2021gsm8k}
Cobbe, K., Kosaraju, V., Bavarian, M., Chen, M., Jun, H., Kaiser, L., Plappert,
  M., Tworek, J., Hilton, J., Nakano, R., et~al.
\newblock Training verifiers to solve math word problems, 2021.
\newblock \emph{arXiv preprint arXiv:2110.14168}, 2021.

\bibitem[Eldan \& Li(2023)Eldan and Li]{eldan2023tinystories}
Eldan, R. and Li, Y.
\newblock Tinystories: How small can language models be and still speak
  coherent english?
\newblock \emph{arXiv preprint arXiv:2305.07759}, 2023.

\bibitem[Gao et~al.(2024)Gao, Qian, Ni, Gan, Hasegawa-Johnson, Chang, and
  Zhang]{gaospeech}
Gao, H., Qian, K., Ni, J., Gan, C., Hasegawa-Johnson, M.~A., Chang, S., and
  Zhang, Y.
\newblock Speech self-supervised learning using diffusion model synthetic data.
\newblock In \emph{International Conference on Machine Learning}, 2024.

\bibitem[Gunasekar et~al.(2023)Gunasekar, Zhang, Aneja, Mendes, Del~Giorno,
  Gopi, Javaheripi, Kauffmann, de~Rosa, Saarikivi,
  et~al.]{gunasekar2023textbooks}
Gunasekar, S., Zhang, Y., Aneja, J., Mendes, C. C.~T., Del~Giorno, A., Gopi,
  S., Javaheripi, M., Kauffmann, P., de~Rosa, G., Saarikivi, O., et~al.
\newblock Textbooks are all you need.
\newblock \emph{arXiv preprint arXiv:2306.11644}, 2023.

\bibitem[Haluptzok et~al.(2023)Haluptzok, Bowers, and Kalai]{haluptzoklanguage}
Haluptzok, P., Bowers, M., and Kalai, A.~T.
\newblock Language models can teach themselves to program better.
\newblock In \emph{International Conference on Learning Representations}, 2023.

\bibitem[Hand(2018)]{mclachlan1988mixture}
Hand, D.~J.
\newblock {Mixture Models: Inference and Applications to Clustering}.
\newblock \emph{Journal of the Royal Statistical Society Series C: Applied
  Statistics}, 2018.

\bibitem[Joshi et~al.(2017)Joshi, Choi, Weld, and
  Zettlemoyer]{joshi2017triviaqa}
Joshi, M., Choi, E., Weld, D.~S., and Zettlemoyer, L.
\newblock Triviaqa: A large scale distantly supervised challenge dataset for
  reading comprehension.
\newblock \emph{arXiv preprint arXiv:1705.03551}, 2017.

\bibitem[Koncel-Kedziorski et~al.(2016)Koncel-Kedziorski, Roy, Amini, Kushman,
  and Hajishirzi]{koncel2016mawps}
Koncel-Kedziorski, R., Roy, S., Amini, A., Kushman, N., and Hajishirzi, H.
\newblock Mawps: A math word problem repository.
\newblock In \emph{North {A}merican Chapter of the Association for
  Computational Linguistics: Human Language Technologies}, 2016.

\bibitem[Lee et~al.(2024)Lee, Wattanawong, Kim, Mangalam, Shen, Anumanchipalli,
  Mahoney, Keutzer, and Gholami]{lee2024llm2llm}
Lee, N., Wattanawong, T., Kim, S., Mangalam, K., Shen, S., Anumanchipalli,
  G.~K., Mahoney, M.~W., Keutzer, K., and Gholami, A.
\newblock Llm2llm: Boosting llms with novel iterative data enhancement.
\newblock In \emph{Annual Meeting of the Association for Computational
  Linguistics}, 2024.

\bibitem[Lester et~al.(2021)Lester, Al-Rfou, and Constant]{lester2021power}
Lester, B., Al-Rfou, R., and Constant, N.
\newblock The power of scale for parameter-efficient prompt tuning.
\newblock In Moens, M.-F., Huang, X., Specia, L., and Yih, S. W.-t. (eds.),
  \emph{Empirical Methods in Natural Language Processing}, 2021.

\bibitem[Li et~al.(2024{\natexlab{a}})Li, Dong, Tang, Wang, Zhang, Huang,
  Huang, Huang, Huang, Zhang, et~al.]{li2024synthetic}
Li, H., Dong, Q., Tang, Z., Wang, C., Zhang, X., Huang, H., Huang, S., Huang,
  X., Huang, Z., Zhang, D., et~al.
\newblock Synthetic data (almost) from scratch: Generalized instruction tuning
  for language models.
\newblock \emph{arXiv preprint arXiv:2402.13064}, 2024{\natexlab{a}}.

\bibitem[Li \& Liang(2021)Li and Liang]{li2021prefix}
Li, X.~L. and Liang, P.
\newblock Prefix-tuning: Optimizing continuous prompts for generation.
\newblock \emph{arXiv preprint arXiv:2101.00190}, 2021.

\bibitem[Li et~al.(2023)Li, Bubeck, Eldan, Del~Giorno, Gunasekar, and
  Lee]{li2023textbooks}
Li, Y., Bubeck, S., Eldan, R., Del~Giorno, A., Gunasekar, S., and Lee, Y.~T.
\newblock Textbooks are all you need ii: phi-1.5 technical report.
\newblock \emph{arXiv preprint arXiv:2309.05463}, 2023.

\bibitem[Li et~al.(2024{\natexlab{b}})Li, Ma, Lu, Lee, Liu, and Guo]{limend}
Li, Y., Ma, X., Lu, S., Lee, K., Liu, X., and Guo, C.
\newblock {MEND}: Meta demonstration distillation for efficient and effective
  in-context learning.
\newblock In \emph{International Conference on Learning Representations},
  2024{\natexlab{b}}.

\bibitem[Liu et~al.(2023)Liu, Bubeck, Eldan, Kulkarni, Li, Nguyen, Ward, and
  Zhang]{liu2023tinygsm}
Liu, B., Bubeck, S., Eldan, R., Kulkarni, J., Li, Y., Nguyen, A., Ward, R., and
  Zhang, Y.
\newblock Tinygsm: achieving> 80\% on gsm8k with small language models.
\newblock \emph{arXiv preprint arXiv:2312.09241}, 2023.

\bibitem[Luo et~al.(2024)Luo, Xu, Zhao, Sun, Geng, Hu, Tao, Ma, Lin, and
  Jiang]{luowizardcoder}
Luo, Z., Xu, C., Zhao, P., Sun, Q., Geng, X., Hu, W., Tao, C., Ma, J., Lin, Q.,
  and Jiang, D.
\newblock Wizardcoder: Empowering code large language models with
  evol-instruct.
\newblock In \emph{International Conference on Learning Representations}, 2024.

\bibitem[Madaan et~al.(2023)Madaan, Tandon, Gupta, Hallinan, Gao, Wiegreffe,
  Alon, Dziri, Prabhumoye, Yang, et~al.]{madaan2024self}
Madaan, A., Tandon, N., Gupta, P., Hallinan, S., Gao, L., Wiegreffe, S., Alon,
  U., Dziri, N., Prabhumoye, S., Yang, Y., et~al.
\newblock Self-refine: Iterative refinement with self-feedback.
\newblock In \emph{Neural Information Processing Systems}, 2023.

\bibitem[Mahajan et~al.(2024)Mahajan, Rahman, Yi, and
  Sigal]{mahajan2024prompting}
Mahajan, S., Rahman, T., Yi, K., and Sigal, L.
\newblock Prompting hard or hardly prompting: Prompt inversion for
  text-to-image diffusion models.
\newblock In \emph{Computer Vision and Pattern Recognition}, 2024.

\bibitem[Mitra et~al.(2023)Mitra, Del~Corro, Mahajan, Codas, Simoes, Agarwal,
  Chen, Razdaibiedina, Jones, Aggarwal, et~al.]{mitra2023orca}
Mitra, A., Del~Corro, L., Mahajan, S., Codas, A., Simoes, C., Agarwal, S.,
  Chen, X., Razdaibiedina, A., Jones, E., Aggarwal, K., et~al.
\newblock Orca 2: Teaching small language models how to reason.
\newblock \emph{arXiv preprint arXiv:2311.11045}, 2023.

\bibitem[Mu et~al.(2023)Mu, Li, and Goodman]{mu2024learning}
Mu, J., Li, X.~L., and Goodman, N.
\newblock Learning to compress prompts with gist tokens.
\newblock In \emph{Neural Information Processing Systems}, 2023.

\bibitem[Mukherjee et~al.(2023)Mukherjee, Mitra, Jawahar, Agarwal, Palangi, and
  Awadallah]{mukherjee2023orca}
Mukherjee, S., Mitra, A., Jawahar, G., Agarwal, S., Palangi, H., and Awadallah,
  A.
\newblock Orca: Progressive learning from complex explanation traces of gpt-4.
\newblock \emph{arXiv preprint arXiv:2306.02707}, 2023.

\bibitem[Pedregosa et~al.(2011)Pedregosa, Varoquaux, Gramfort, Michel, Thirion,
  Grisel, Blondel, Prettenhofer, Weiss, Dubourg, et~al.]{scikit-learn}
Pedregosa, F., Varoquaux, G., Gramfort, A., Michel, V., Thirion, B., Grisel,
  O., Blondel, M., Prettenhofer, P., Weiss, R., Dubourg, V., et~al.
\newblock Scikit-learn: Machine learning in python.
\newblock \emph{Journal of Machine Learning Research}, 2011.

\bibitem[Petrov et~al.(2024)Petrov, Torr, and Bibi]{petrovprompting}
Petrov, A., Torr, P., and Bibi, A.
\newblock When do prompting and prefix-tuning work? a theory of capabilities
  and limitations.
\newblock In \emph{International Conference on Learning Representations}, 2024.

\bibitem[Pillutla et~al.(2021)Pillutla, Swayamdipta, Zellers, Thickstun,
  Welleck, Choi, and Harchaoui]{pillutla2021mauvemeasuringgapneural}
Pillutla, K., Swayamdipta, S., Zellers, R., Thickstun, J., Welleck, S., Choi,
  Y., and Harchaoui, Z.
\newblock {MAUVE}: Measuring the gap between neural text and human text using
  divergence frontiers.
\newblock In \emph{Neural Information Processing Systems}, 2021.

\bibitem[Shi et~al.(2023)Shi, Chen, Misra, Scales, Dohan, hsin Chi, Scharli,
  and Zhou]{shi2023largelanguagemodelseasily}
Shi, F., Chen, X., Misra, K., Scales, N., Dohan, D., hsin Chi, E.~H., Scharli,
  N., and Zhou, D.
\newblock Large language models can be easily distracted by irrelevant context.
\newblock In \emph{International Conference on Machine Learning}, 2023.

\bibitem[Sun et~al.(2023)Sun, Chen, Wang, Li, and Gao]{sun2023transcoder}
Sun, Q., Chen, N., Wang, J., Li, X., and Gao, M.
\newblock Transcoder: Towards unified transferable code representation learning
  inspired by human skills.
\newblock \emph{arXiv preprint arXiv:2306.07285}, 2023.

\bibitem[Team et~al.(2024)Team, Riviere, Pathak, Sessa, Hardin, Bhupatiraju,
  Hussenot, Mesnard, Shahriari, Ram{\'e}, et~al.]{team2024gemma}
Team, G., Riviere, M., Pathak, S., Sessa, P.~G., Hardin, C., Bhupatiraju, S.,
  Hussenot, L., Mesnard, T., Shahriari, B., Ram{\'e}, A., et~al.
\newblock Gemma 2: Improving open language models at a practical size.
\newblock \emph{arXiv preprint arXiv:2408.00118}, 2024.

\bibitem[Yu et~al.(2024)Yu, Jiang, Shi, Jincheng, Liu, Zhang, Kwok, Li, Weller,
  and Liu]{yumetamath}
Yu, L., Jiang, W., Shi, H., Jincheng, Y., Liu, Z., Zhang, Y., Kwok, J., Li, Z.,
  Weller, A., and Liu, W.
\newblock Metamath: Bootstrap your own mathematical questions for large
  language models.
\newblock In \emph{International Conference on Learning Representations}, 2024.

\bibitem[Zhang et~al.(2023)Zhang, Rao, and Agrawala]{zhang2023adding}
Zhang, L., Rao, A., and Agrawala, M.
\newblock Adding conditional control to text-to-image diffusion models.
\newblock In \emph{International Conference on Computer Vision}, 2023.

\end{thebibliography}
\bibliographystyle{icml2025}

\clearpage
\appendix

\section{Empirical Evaluation Additional Details}
\label{sec:empirical_appendix}
This appendix provides supplementary details about our empirical evaluations.

\subsection{Benchmark Task Discussion}
\label{app:task_discussion}
We consider the following tasks:

  {\bf{Code -- MBPP \citep{austin2021program}.}} For the coding domain, we consider the ``Mostly Basic Python Problems'' (MBPP) benchmark, which consists of short Python programming exercises, e.g. ``Write a python function to find the first repeated character in a given string'' and answers written in Python code. The task is evaluated using a 3-shot prompt (i.e. a prompt pre-pended with three instructional examples) and pass@1 metric, measuring if a single top generated result is correct.
  
  {\bf{Math -- GSM8K \citep{cobbe2021gsm8k}.}} For the math domain, we use the grade school math word problems from the GSM8K benchmark. This dataset contains only highly-curated word problems written by humans that are conceptually simple, but require multi-step reasoning. 
  For this generative task we use a 5-shot prompt, again, measuring the pass@1 metric.
  
  {\bf{Reasoning -- BoolQ \citep{clark2019boolq}.}} For general reasoning, we consider the binary question answering dataset from the BoolQ benchmark. The questions arise organically from anonymized Google search queries which can be answered as either `true' or `false'.
  Each question and answer is paired with a passage (average length of 108 tokens) that is extracted from a relevant Wikipedia page. This is evaluated as a scoring/classification task and accuracy is reported.

The MBBP task requires basic Python programming knowledge to solve, but the questions are generally short, follow a similar pattern and are concerned with a relatively narrow set of themes.  The GSM8K task requires basic math and language comprehension skills, but generating the problem is arguably even harder than solving it. It requires generating a premise containing several numerical quantities and then a question that can be answered using the provided information in a non-trivial fashion. Nonetheless the premises are somewhat formulaic and thematically similar.  

The BoolQ reasoning task, which requires general reading comprehension to solve, is perhaps the most difficult task to generate synthetic data for.  Generating a problem requires writing a long (relative to MBPP and GSM8K) passage on an arbitrary topic that contains a collection of facts, but that does not necessarily stick to any formula or theme, and then generate a true/false question that can be answered directly by the passage. 
As we shall see in the empirical evaluation, the level of difficulty in generating a high-quality question can impact the relative quality and value of generated synthetic data.


\subsection{Details on ``Step 2. Generate Questions''}\label{app:postprocess}

In the case of SoftSRV, we cycle through questions found in the training fold of the benchmark dataset, $x_i$, and produce a new synthetic questions $x_i' \sim H(\Pb_\theta(\emb(x_i)))$ via temperature sampling (with default temp=1). For the $\FSP$ variant, no context vector and, thus, no training examples are used during generation. For all SoftSRV methods, no natural language prompt template of any kind is used. 

For the baselines, we cycle through questions found in the training fold and insert them into the relevant domain-specific prompt template
and generate synthetic questions by querying the same backbone LLM used by SoftSRV. We conducted a search over temperature=\{1,2,4\} and found a temperature of 2 to provide a balance of diversity and quality for the baseline approaches.

For all methods, we generate 100K questions, repeating example questions from the training fold in a round-robin fashion. We then run a simple filtering, deduplication and subsampling pipeline to arrive at a target fine-tuning dataset size $N_s$.  Concretely, from these 100K, we first filter exact duplicates.  Then, to encourage a diverse subsample, again for all methods, we cluster the data and select examples from each cluster randomly in round-robin fashion. That is, using the scikit-learn library \citep{scikit-learn}, we apply MiniBatch $k$-means to vectorized data, which has been reduced in dimensionality using SVD. For all methods, we set the number of clusters for MiniBatch $k$-means to 700, reduced the dimensionality to 100 features and used sk.TfidfVectorizer for vectorization.  Given the $k$-means clustering, we randomly select without replacement one point per cluster  until $N_s$ questions are chosen.  We use $N_s$=50,000 for MBPP and GSM8K and $N_s$=20,000 for BoolQ.

\subsection{Decontamination Process} \label{app:decontaminate}

Even though the test set was never used in our synthetic data generation pipeline, the frozen LLM models that are leveraged to generate questions and answers might have been exposed to the test set during their pretraining phase. Thus, for all methods, we decontaminate the generated sequences against the respective benchmark's test set by removing any n-gram matches where $n=13$ as is common practice \citep{Brown2020}. Prior to calculating the matches, we eliminate all punctuation and numerical characters. We found that the contamination of the generated sequences to the test set is minimal with less than $0.1\%$ for GSM8K and MBPP and around $1\%$ for BoolQ. 


\subsection{Diversification of the $\HP$ method} \label{app:diversify}
\begin{figure}
    \centering
        \includegraphics[scale=0.5]{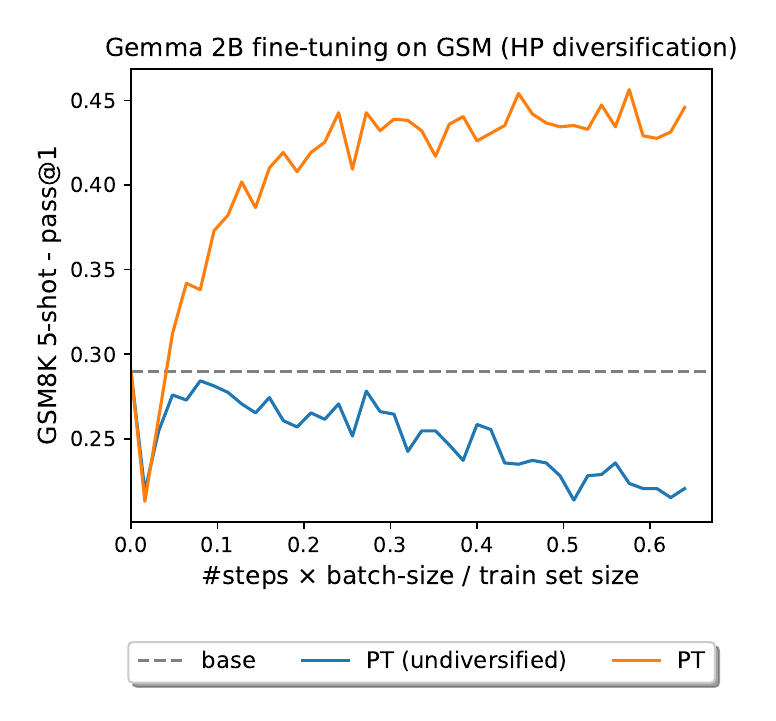}
    \caption{Performance of $\HP$ method with and without diversification, i.e. the phrase generate "10 different questions".}
    \label{fig:hpdiversification}
\end{figure}

To arrive at the the $\HP$ method presented in the main body of the paper, we conducted multiple iterations of prompt engineering and template refinements. In particular, we found that asking the model to generate "10 different questions" per example question and using a higher decoding temperature was critical. 
We demonstrate this effect on the GSM8K benchmark in 
Figure~\ref{fig:hpdiversification} which compares the performance of a model fine-tuned on data generated by $\HP$ method to that of the model fine-tuned on its undiversified counterpart where the question template asks to generate one question per given example question and the default decoding temperature is used. We found similar results for the other datasets. Despite the relative small change in the settings, the difference in model performance is significant, demonstrating some of the idiosyncratic nature of prompt template approaches.  We provide the template for the undiversified $\HP$ in Appendix~\ref{app:hp}.

\subsection{Comparison with Curated Prompt Template Generated Dataset (TinyGSM)}\label{tinygsm}
 \begin{figure}
  \centering
  \includegraphics[scale=0.5]{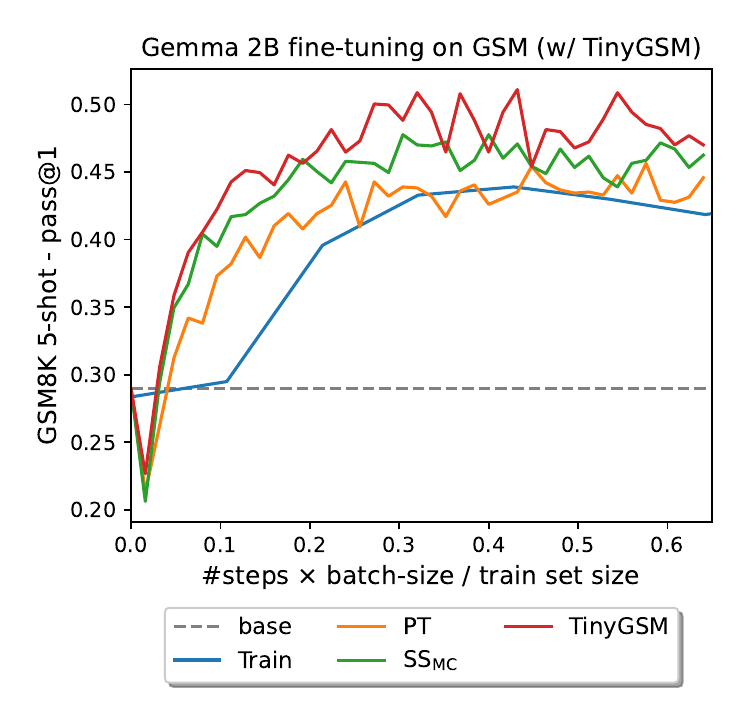}
  \caption{For the GSM8K benchmark, we show the Gemma 2B performance after fine-tuning on the $\MLPC$ and $\HP$ generated datasets against a sample of the same size from the curated TinyGSM dataset.} 
  \label{fig:gsm-modern}
\end{figure}


Here, we compare against a high quality synthetic dataset, TinyGSM, within our fine-tuning evaluation framework. TinyGSM was expertly curated by \cite{liu2023tinygsm} for GSM8K-PAL, which is a program aided language model (PAL) variant of GSM8K that asks for questions to be answered in the form of Python functions. This has the advantage of enabling verification of the answer in a programmatic fashion.
\cite{liu2023tinygsm} use GPT-3.5-turbo with prompt templates seeded with training questions from the original GSM8K dataset and from the GSM-IC dataset, which is a dataset crafted to incorporate irrelevant context in order to bolster model robustness \citep{shi2023largelanguagemodelseasily}. They use two types of prompts: the first asks to generate both questions and answers while the second requires two calls to the LLM to first generate a question and then an answer. Leveraging the fact that the solutions of the math word problems are written in Python, they then filter out any data that contains code that is not executable by a Python interpreter. They additionally filter out questions that do not contain numbers as this indicates flawed math problems.

In order to evaluate the TinyGSM generated question in our setting, we randomly sample 100K questions from the publicly available TinyGSM dataset, then further subsample down to 50K using the same post-processing pipeline used by all other methods in our comparison. Finally, we generate answers, fine-tune and evaluate in the same fashion as the other prompt template baselines.
Figure~\ref{fig:gsm-modern} shows that a model fine-tuned on the $\MLPC$ dataset attains a performance close to that when fine-tuned on the sample from the $\TGSM$ dataset while the $\HP$ method lags behind both.  TinyGSM performing closely to $\MLPC$ is encouraging given that the TinyGSM dataset is highly curated and tailored specifically for the GSM8K benchmark. 

 \subsection{Mixture of Prompts with Various Values of $k$}\label{app:mixture}

\begin{figure}
    \centering
    \includegraphics[scale=0.5]{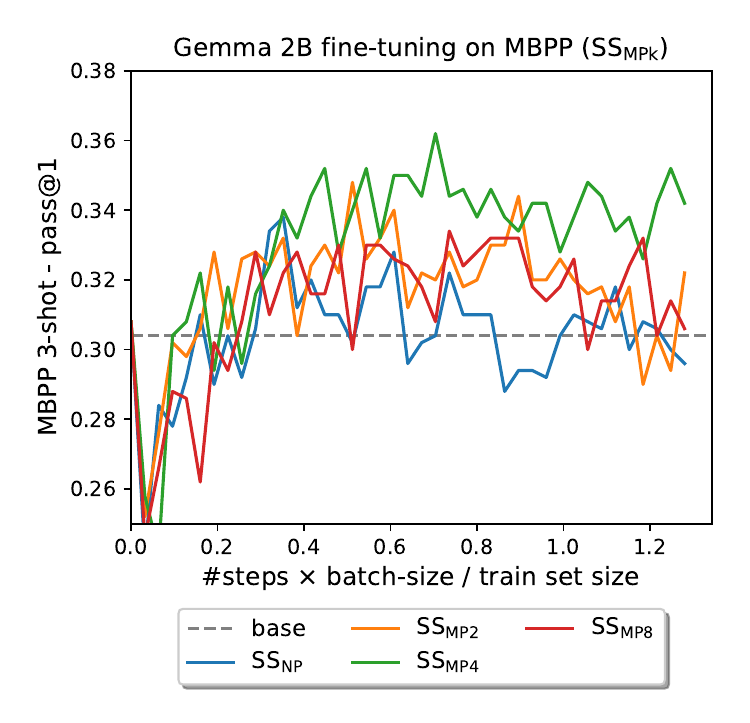}
    \includegraphics[scale=0.5]{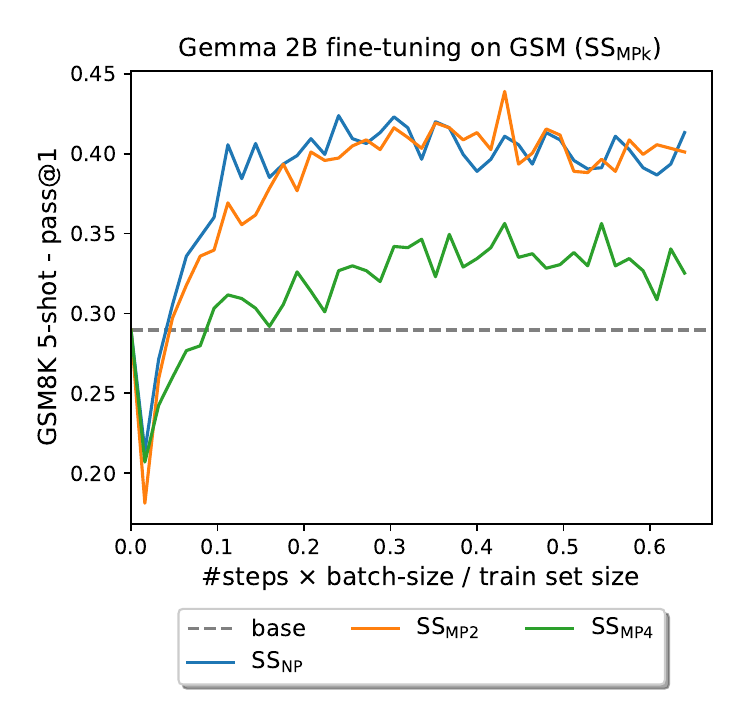}
    \includegraphics[scale=0.5]{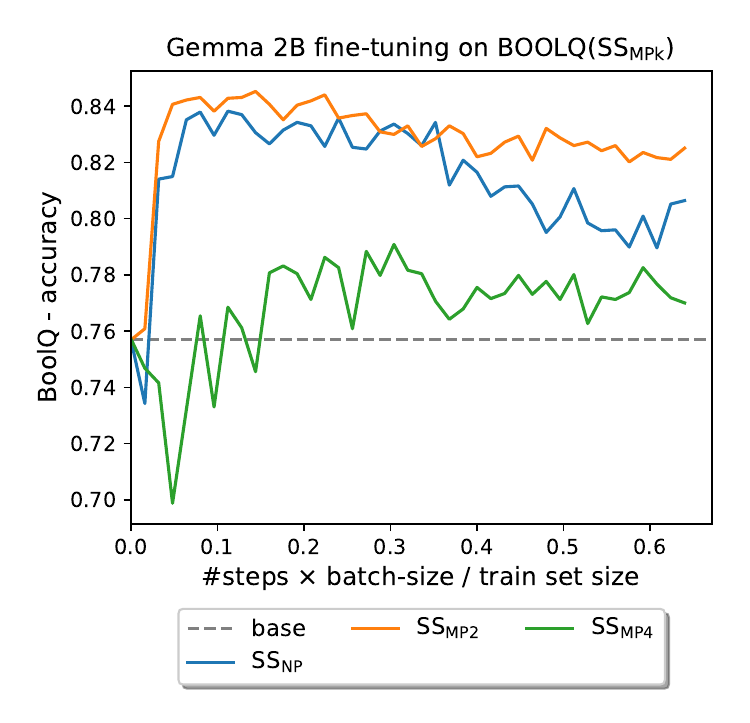}
    \caption{Comparison of $\MPK$ variants for different values of $k$. $\FSP$ corresponds to $k=1$.}
    \label{fig:mpkplot}
\end{figure}

Here, we conduct a exploration to measure the effect of changing the number of basis embedding matrices, $k$, of the $\MPK$ method.

Figure \ref{fig:mpkplot} shows the comparison in performance on various benchmarks across different values of $k$. We see an inherent trade-off as we increase $k$, which increases the capacity of the contextual embedding but also then require training more parameters. For MBPP we see that the value $k=4$ appears to be optimal, while for GSM8K and BOOLQ we see performance peaks at $k=2$. We fix $k=2$ throughout the evaluations in the main paper as a good general choice across different tasks.

\subsection{{\sc Mauve} Score Computation}\label{app:mauve}
We let $\mathcal{G}$ and $\mathcal{D}$ denote the generated and target/reference distributions, respectively, and compute the {\sc mauve} scores for each synthetic dataset as follows. Using an embedding model, a vector representation is computed for each sequence in the synthetic and reference sets. These embeddings are then projected into a discrete set using $k$-means clustering, and a divergence curve is traced between the cluster distributions of $\mathcal{G}$ and $\mathcal{D}$, see Equation~1 in \cite{pillutla2021mauvemeasuringgapneural}. The {\sc mauve} has value between 0 and 1 and corresponds to the area under the divergence curve, where higher scores are indicative of a closer match between $\mathcal{G}$ and $\mathcal{D}$.
The same small LM serving as the context embedder $E$ for SoftSRV  in  Subsection~\ref{pipeline} is used here to compute per-token representations, which is then averaged to produce the sequence-level embedding. We use $k=32$ clusters for all domains as we found that $k=16$ or $k=64$ yields similar qualitative results.

In Table~\ref{mauve-table-full} we additionally report the {\sc mauve} score computed when using the training set as a reference sample.

\begin{table*} 
    \centering
    \caption{The {\sc mauve} similarity scores of synthetic datasets computed with respect to the non-synthetic training and test fold of each dataset. The similarity computed between the train and test fold itself is also included in the final row of the table.}
    \vspace{0.1in}
    \begin{tabular}{|l|c|c|c|c|c|c|} 
        \hline
        & \multicolumn{3}{c|}{Train} & \multicolumn{3}{c|}{Test} \\
        & MBPP & GSM8K & BoolQ & MBPP & GSM8K & BoolQ \\
        \hline
        $\HP$ & 0.622 & 0.933  & 0.663 & 0.463 & 0.914 & 0.784 \\
        $\HPSR$ & 0.327  & 0.870 & --  & 0.397 & 0.865 & -- \\
        $\FSP$ &   0.776  & 0.862 & 0.519 &   0.781  & 0.839 & 0.575 \\
        $\MPtwo$ & 0.722 & 0.727 & 0.731 & 0.646 &0.735 & 0.689 \\
        $\MLPC$  & 0.604 &  0.993  &   0.997  & 0.477 &   0.991  &   0.995   \\
        \hline
        Train &  1.000 & 1.000 & 1.000 & 0.963 & 0.998 & 0.999 \\
        \hline
    \end{tabular}
    \label{mauve-table-full}
\end{table*}

\section{Related Work on Prompt-tuning \& Non-text Data Modalities}\label{app:soft_prompt_related_work}
 
Prompt-tuning is primarily known for its use as a parameter efficient fine-tuning method \citep{lester2021power,li2021prefix}. It has also been used as a framework to learn or compress in-context instructions \citep{mu2024learning}. Recently, \citet{limend} proposed training a secondary model to compress in-context instructions into soft prompt that is then prepended to the task prompt templates. 

While our study has been focused on generating synthetic text, there have been similar efforts in other modalities. For example, the ControlNet approach of \citet{zhang2023adding} trains a diffusion model to produce images conditioned on contextual input, for example, image edges or 3D pose skeletons. Similarly, \citet{gaospeech} trains a diffusion-based speech model to condition on a ``simple speech representation'' embedding to guide the generation of new synthetic speech data. Finally, in the case of text-to-image generators, there has been a significant amount of work in solving the ``inverse'' problem of mapping from an image back to a prompt, so that one can  more predictably generate synthetic images in certain styles (see \cite{mahajan2024prompting} and many references therein).

\section{Natural Language Prompt Templates}\label{app:hp}

Below, we report the templates used for prompt template baselines for each benchmark dataset. Figure~\ref{fig:question} provides the question templates while Figure~\ref{fig:answer} shows the answer templates. Figure~\ref{fig:undiversified} provides the template for the undiversified $\HP$ method, described in Appendix~\ref{app:diversify}, for the GSM8K benchmark. Figure~\ref{fig:refine} reports the critique and refine templates for $\HPSR$. 

\section{Generated Text Examples}
\label{app:examples}

In Figure~\ref{fig:generated_text_examples}, we provide examples of synthetic questions for GSM8K (math questions), MBPP (Python questions) and BoolQ (passage and question pairs) generated using the $\MLPC$ method. Notice that the questions generated for BoolQ require longer range dependencies and may be more complex to generate and answer. Finally, for comparison, in Figure~\ref{fig:train_set_examples}, we show examples from the train split of each dataset. These train set questions appear similar in style to the questions generated by the $\MLPC$ method.

\begin{figure}
 \centering
 \textbf{ MBPP Question Template:}
    \fbox {
    \parbox{0.98\linewidth}{ \small {\texttt{Consider the following python question: \smallbreak
     \hspace{1cm} [insert example question] \smallbreak
 Now generate 10 different questions that require writing a Python function similar to the example above. Make sure each question is different and sufficiently rephrased. Please make sure you generate questions, and not answers. Please make sure each question you generate has a well-defined answer. \smallbreak
      Question 1: }}}}
      
      \vspace{5mm}
    \textbf{GSM8K Question Template:}
      \fbox {
    \parbox{0.95\linewidth}{ \small {\texttt{Consider the following grade-school math problem: \smallbreak
     \hspace{1cm} [insert example question] \smallbreak
      Now generate 10 different questions that require solving a grade-school math problem similar to the example above. Make sure each question is different and sufficiently rephrased. Please make sure you generate questions, and not answers. Please make sure each question you generate has a well-defined answer.\smallbreak
      Question 1:}}}}
      
      \vspace{5mm}
      \textbf{BoolQ Question Template:}
     \fbox {
    \parbox{0.98\linewidth}{ \small \texttt{Consider the following passage and question:\smallbreak
     \hspace{1cm} [insert example question] \smallbreak
    Now generate 10 different passages and questions similar to the example above. Please make sure each question you generate has a boolean answer that can be answered by the passage. Make sure each passage and question is different and sufficiently rephrased.  Please make sure you generate passages and questions, and not answers.\smallbreak
    Passage and Question 1:}}}
      \caption{Question template for MBPP, GSM8K and BoolQ benchmarks for the $\HP$ method.}
          \label{fig:question}
\end{figure}

  \begin{figure}
    \centering
      \textbf{ MBPP Answer Template:}

 \fbox {
    \parbox{0.98\linewidth}{ \small\texttt{Please answer the following python question:
\smallbreak
     \hspace{1cm} [insert example question] \smallbreak
      Please generate your answer as a Python function. The docstring of the function should contain the above question as-is, without any modification. Please make sure that your function is valid Python code that compiles. Please try your best to correctly answer the question.  \smallbreak
      Answer:}}}
   
   \vspace{5mm}
     \textbf{ GSM8K Answer Template:}
   \fbox {
    \parbox{0.95\linewidth}{ \small \texttt{Please answer the following question that tests reasoning:
     \smallbreak \hspace{1cm} [insert example question] \smallbreak
      Answer:
  } }}
  
  \vspace{5mm}
      \textbf{BoolQ Answer Template:}
   \fbox {
    \parbox{0.98\linewidth}{ \small   \texttt{Please answer the following question based on the passage. Your answer should be either True or False. Do not provide any other justification.    \smallbreak \hspace{1cm} [insert example question] \smallbreak
      Answer:
  }}}
  \caption{Answer template for MBPP, GSM8K and BoolQ benchmarks for the $\HP$ and $\HPSR$ method.}
    \label{fig:answer}
  \end{figure}
  \begin{figure}
    \centering
 \textbf{ GSM8K Question Template for undiversified $\HP$:}
    \fbox {  \parbox{0.95\linewidth}{ \small\texttt{
 Please generate a question that requires solving a grade-school math problem. Here is an example of such a question:
 \smallbreak
     \hspace{1cm} [insert example question] \smallbreak
      Now generate a new question. Please make sure your question is not too similar to the example above. Please make sure you generate a question, and not an answer. Please make sure the question you generate has a well-defined answer.
}}}
\caption{Question template for the undiversified $\HP$ method for the GSM8K benchmark. }
\label{fig:undiversified}
\end{figure}

\begin{figure}
\centering
\textbf{Critique Template:}

     \fbox {
    \parbox{0.95\linewidth}{ \small \texttt{Please provide actionable feedback on the clarity, difficulty, and originality of the following \{Python question, grade school math problem, passage/question problem\}:
      \smallbreak \hspace{1cm} [insert question] \smallbreak
}}}
 
\vspace{5mm}
\textbf{MBPP Refine Template:}

\fbox {
    \parbox{0.98\linewidth}{ \small
      \texttt{Read the following Python question and the critique, and write a new Python question based on the critique:\smallbreak
    Question: 
      \smallbreak \hspace{1cm} [insert question] \smallbreak
    Critique: 
      \smallbreak \hspace{1cm} [insert critique] \smallbreak
    If the critique is strongly positive, say 'Stop'.
    Otherwise, write a new Python question in a single sentence starting with 'Write a Python function' based on the critique.
    Do not ask for docstring or test cases.
}}}

\vspace{5mm}
\textbf{GSM8K Refine Template:}
    \fbox {
    \parbox{0.98\linewidth}{ \small  \texttt{Read the following grade-school math problem and the critique, and write a new grade-school math problem based on the critique:\smallbreak
    Question: 
      \smallbreak \hspace{1cm} [insert question] \smallbreak
    Critique: 
      \smallbreak \hspace{1cm} [insert critique] \smallbreak
    If the critique is strongly positive, say 'Stop'.
    Otherwise, write a new grade-school math problem based on the critique. 
    Write the question only, do not include the answer. 
}}}

\vspace{5mm}
\textbf{BoolQ Refine Template:}
\fbox {
    \parbox{0.98\linewidth}{ \small\texttt{Read the following passage/question problem and the critique, and write a new passage/question problem based on the critique:\smallbreak
    Question: 
     \smallbreak \hspace{1cm}  [insert question/passage]  \smallbreak
    Critique:
     \smallbreak \hspace{1cm}  [insert critique]  \smallbreak
    If the critique is strongly positive, say 'Stop'.
    Otherwise, write a new passage/question problem based on the critique.
    Write the passage and question only, do not include the answer.
} }}
  \caption{Refine template for MBPP, GSM8K and BoolQ benchmarks for the $\HPSR$ method.}
  \label{fig:refine}
  \end{figure}

 \DefineVerbatimEnvironment%
{VerbatimExamples}{Verbatim}
{numbers=none,fontsize=\scriptsize,frame=single,framerule=0.1mm}

\begin{figure}
\centering
\textbf{GSM8K Generated Text Examples}
\begin{VerbatimExamples}
Julian is twice as old as his little brother, Eli. In
12 years, Julian will be 37. How old is Eli today?
\end{VerbatimExamples}
\begin{VerbatimExamples}
Sarah used 1/3 of her grass cutting saw's fuel in
cutting the grass in front of her house and used 2/5
of the remaining fuel to cut the grass in her back
yard. If she started with 30 liters of fuel, how much
fuel did she use in cutting the front yard?
\end{VerbatimExamples}
\begin{VerbatimExamples}
A boy is jogging at 4 m/s. How long will he take to
cover a distance of 5 km?
\end{VerbatimExamples}

\textbf{MBPP Generated Text Examples}
\begin{VerbatimExamples}
Write a python function to find the average of three
given numbers.
\end{VerbatimExamples}
\begin{VerbatimExamples}
Given a sequence of characters, create a new sequence
such that every character is repeated k times.
\end{VerbatimExamples}
\begin{VerbatimExamples}
Write a function that takes in two numbers and returns
the sum of the two numbers.
\end{VerbatimExamples}

\textbf{BoolQ Generated Text Examples}
\begin{VerbatimExamples}
PASSAGE: There are no known cases of anyone catching
a common cold by exposure to the cold weather. The
most likely cause of a common cold is a virus. The
symptoms of the common cold can include a runny nose,
sore throat, cough, sneezing, nasal congestion, and
watery eyes. A common cold is considered to be mild
when only mild symptoms are present. Severe symptoms
include sinus pain and fever.
QUESTION: can the common cold be caused by cold
weather
\end{VerbatimExamples}
\begin{VerbatimExamples}
PASSAGE: Land Before Time V: The Mysterious Island
(1997) is an American animated direct-to-video film. 
It is the fifth film in The Land Before Time series
and the first film in the series to be animated in
Canada and also the first film in the series to be
created by Universal Cartoon Studios.
QUESTION: is the land before time 5 a sequel
\end{VerbatimExamples}
\begin{VerbatimExamples}
PASSAGE: The semilunar valves of the heart (or 
sometimes called the sigmoid valves) are the valves
that prevent backflow from the aorta and pulmonary
artery. They are situated at the beginning of these
arteries. The pulmonary semilunar valve is on the
right side of the heart, while the aortic semilunar
valve is on the left. They both have three leaflets,
each of which is attached to the arterial wall by the
corpora fibrosa. They are attached to the artery in
such a manner that they cannot fold over to block the
flow of blood.
QUESTION: are the aortic and pulmonary valves
semilunar valves
\end{VerbatimExamples}

\caption{Examples of generated synthetic questions generated by $\MLPC$.}
\label{fig:generated_text_examples}
\end{figure}

\begin{figure}
\centering
\textbf{GSM8K Example Questions From Train Set}
\begin{VerbatimExamples}
A choir was singing a song that involved 30 singers. 
In the first verse, only half of them sang. In the
second verse, a third of the remaining singers joined
in. How many people joined in the final third verse
that the whole choir sang together?
\end{VerbatimExamples}
\begin{VerbatimExamples}
Kyle is 5 years older than Julian. Julian is 20 years
younger than Frederick. Frederick is 2 times older
than Tyson. If Tyson is 20, how old is Kyle?
\end{VerbatimExamples}
\begin{VerbatimExamples}
Mr. Williams bought 10 gallons of juice for a party.
Each gallon has 10 cups. At the party, 5 cups of juice
were left. How many cups of juice were drunk?
\end{VerbatimExamples}

\textbf{MBPP Example Questions From Train Set}
\begin{VerbatimExamples}
Write a function to move all the numbers in it to the
given string.
\end{VerbatimExamples}
\begin{VerbatimExamples}
Write a function to find the largest subset where each
pair is divisible.
\end{VerbatimExamples}
\begin{VerbatimExamples}
Write a function to increment the numeric values in
the given strings by k.
\end{VerbatimExamples}

\textbf{BoolQ Example Questions From Train Set}
\begin{VerbatimExamples}
PASSAGE: Since being fixed on the fourth Thursday in
November by law in 1941, the holiday in the United
States can occur on any date from November 22 to 28. 
When it falls on November 22 or 23, it is not the 
last Thursday, but the penultimate Thursday in 
November. Regardless, it is the Thursday preceding
the last Saturday of November.
QUESTION: is thanksgiving the last thursday of 
november every year
\end{VerbatimExamples}
\begin{VerbatimExamples}
PASSAGE: The Rocky Mountains, also known as the
Rockies, are a major mountain range in western North
America. The Rocky Mountains stretch more than 3,000
miles (4,800 km) from the northernmost part of British
Columbia, in western Canada, to New Mexico, in the 
Southwestern United States. Within the North American
Cordillera, the Rockies are somewhat distinct from the
Pacific Coast Ranges, Cascade Range, and the Sierra
Nevada, which all lie further to the west.
QUESTION: is the sierra nevada part of the rocky
mountains
\end{VerbatimExamples}
\begin{VerbatimExamples}
PASSAGE: The first player to get rid of their last
card (``going out'') wins the hand and scores points
for the cards held by the other players. Number cards
count their face value, all action cards count 20, and
Wild and Wild Draw Four cards count 50. If a Draw Two
or Wild Draw Four card is played to go out, the next
player in sequence must draw the appropriate number of
cards before the score is tallied.
QUESTION: can you finish a uno game on a wild card
\end{VerbatimExamples}

\caption{Examples of questions from the train sets.}
\label{fig:train_set_examples}
\end{figure}

\end{document}